\documentclass[dvipsnames,format=sigconf,anonymous=false,review=false]{acmart}

\usepackage{stmaryrd}
\usepackage{amsmath}
\usepackage{amsfonts}

\SetSymbolFont{stmry}{bold}{U}{stmry}{m}{n}

\usepackage{ifthen}
\usepackage{algorithm}
\usepackage{algorithmic}

\newcommand{\btheta}{\boldsymbol{\theta}}

\newcommand{\x}{\boldsymbol{x}}
\newcommand{\y}{\boldsymbol{y}}

\newcommand{\R}{\mathbb{R}}
\newcommand{\X}{\mathcal{X}}
\newcommand{\diff}{\mathrm{d}}

\newcommand{\argmin}{\mathop{\rm arg~min}\limits}

\newcommand{\T}{\mathrm{T}}

\usepackage{xcolor}
\usepackage{ifthen}
\newcommand{\markupdraft}[2]{
    \ifthenelse{\equal{#1}{display}}{#2}{}
    \ifthenelse{\equal{#1}{color}}{\color{#2}}{}
}

\newcommand{\newcolored}[3][]{{\markupdraft{color}{#2}#3}
    \ifthenelse{\equal{#1}{}}{}{\markupdraft{display}{{\color{yellow!70!black}[#1]}}}}
\newcommand{\del}[2][]{{\markupdraft{display}{{\color{orange}[removed: ``#2''[#1]]}}}} 
\newcommand{\new}[2][]{\newcolored[#1]{blue}{#2}}

\newcommand{\calc}[2][]{{\markupdraft{display}{{\color{purple} Derived as follows: {#2}[#1]}}}} 

\renewcommand{\del}[2]{}  
\renewcommand{\calc}[2]{}  
\renewcommand{\markupdraft}[2]{}  

\AtBeginDocument{%
  }

\copyrightyear{2026}
\acmYear{2026}
\setcopyright{cc}
\setcctype{by}
\acmConference[GECCO '26]{Genetic and Evolutionary Computation Conference}{July 13--17, 2026}{San Jose, Costa Rica}
\acmBooktitle{Genetic and Evolutionary Computation Conference (GECCO '26), July 13--17, 2026, San Jose, Costa Rica}
\acmDOI{10.1145/3795095.3805125}
\acmISBN{979-8-4007-2487-9/2026/07}

\begin{document}

\title[Adaptive Stochastic Natural Gradient Method for Safe Optimization on Binary Space]{Adaptive Stochastic Natural Gradient Method\\for Safe Optimization on Binary Space}


\author{Kento Uchida}
\email{uchida-kento-fz@ynu.ac.jp}
\orcid{0000-0002-4179-6020}
\affiliation{%
  \institution{Yokohama National University}
  \city{Yokohama}
  \state{Kanagawa}
  \country{Japan}
  \postcode{240-8501}
}

\author{Ryoki Hamano}
\email{hamano_ryoki_xa@cyberagent.co.jp}
\orcid{0000-0002-4425-1683}
\affiliation{%
  \institution{CyberAgent}
  \city{Shibuya}
  \state{Tokyo}
  \country{Japan}
  \postcode{150-0042}
}

\author{Masahiro Nomura}
\email{nomura@comp.isct.ac.jp}
\orcid{0000-0002-4945-5984}
\affiliation{%
  \institution{Institute of Science Tokyo}
  \city{Meguro}
  \state{Tokyo}
  \country{Japan}
  \postcode{152-8550}
}

\author{Shinichi Shirakawa}
\email{shirakawa-shinichi-bg@ynu.ac.jp}
\orcid{0000-0002-4659-6108}
\affiliation{%
  \institution{Yokohama National University}
  \city{Yokohama}
  \state{Kanagawa}
  \country{Japan}
  \postcode{240-8501}
}

\renewcommand{\shortauthors}{K. Uchida et al.}

\begin{abstract}
Optimization problems in real-world applications across the medical and engineering domains often involve potential risks when evaluating candidate solutions.
Safe optimization aims to perform optimization while suppressing unsafe solution evaluations in such situations.
For continuous search spaces, there exist safe optimization methods based on evolutionary computation.
However, the algorithm development of safe optimization methods for binary search spaces has not been adequately addressed.
In this study, we incorporate additional mechanisms for safe optimization into a binary optimization method, the adaptive stochastic natural gradient method (ASNG) with a family of Bernoulli distributions.
For safety functions that must be kept non-negative during optimization, the proposed method, safe ASNG, estimates the Lipschitz constants with respect to the Hamming distance by constructing surrogate models of safety functions based on discrete Walsh functions.
Then, safe ASNG computes a safe region that consists of safe solutions around the previously evaluated safe solutions.
By projecting newly generated solutions to their nearest neighbors within the safe region, safe ASNG suppresses unsafe solution evaluations.
Experimental results on benchmark problems on binary domains confirm that, while the comparative methods fail to suppress unsafe solution evaluations, safe ASNG achieves efficient optimization while effectively suppressing unsafe solution evaluations.
\end{abstract}




\maketitle

\section{Introduction}

In optimization problems arising in medical and engineering domains, there exist problems in which evaluating candidate solutions may involve potential risks \cite{safeopt:application:descrete:1,safeopt:application:2,Louis:2022,Modugno:2016,safeopt:application:1,safeopt:application:descrete:2}.
For example, in metal milling processes, optimization of control scheduling with constraints on the upper bounds of motor output is required in order to prevent machine failures \cite{safeopt:application:1}.
Similarly, in the optimization of drug combinations for improving therapeutic efficacy, optimization is performed under toxicity constraints that take into account patient-specific variations in drug responses \cite{safeopt:application:descrete:2}.
Optimization conducted while avoiding such risky solution evaluations is referred to as safe optimization~\cite{survey}, and it is formulated as a constrained optimization problem that aims to suppress the evaluation of unsafe solutions whose safety function values computed during evaluation fall below a prescribed threshold.

To achieve efficient safe optimization, a variety of optimization methods have been proposed~\cite{safecmaes,Kaji:2009,safeopt}.
SafeOpt~\cite{safeopt} is a Bayesian optimization-based approach that suppresses the evaluation of unsafe solutions by computing a safe region consisting only of solutions whose evaluations are guaranteed to be safe, based on the Lipschitz constant of the safety function given in advance.
In addition, as a generic strategy for safe optimization that can be incorporated into evolutionary computation, violation avoidance~\cite{Kaji:2009} has been proposed, in which solution generation is repeatedly retried until the nearest previously evaluated solution is a safe one.
Evolutionary computation methods generally exhibit superior computational efficiency in updates and better convergence properties compared with Bayesian optimization, and therefore the development of evolutionary algorithms that are effective for safe optimization is highly desired.

As an approach for achieving efficient safe optimization in black-box continuous optimization problems, safe CMA-ES~\cite{safecmaes} has been proposed, which is an optimization method based on the covariance matrix adaptation evolution strategy (CMA-ES)~\cite{hansen:1996:ec}.
Safe CMA-ES estimates the Lipschitz constant of the safety function by exploiting gradient information obtained from Gaussian process regression trained on previously evaluated solutions, and computes a safe region consisting of safe solutions only.
Then, solutions sampled from a multivariate Gaussian distribution are mapped to their nearest neighbors within the safe region to suppress unsafe solution evaluations.
Although safe CMA-ES is an effective method for achieving efficient safe optimization, its applicability is limited to continuous optimization problems.
\new{As shown in the application of drug combinations~\cite{safeopt:application:descrete:2}, there is a demand for safe optimization in the binary domain.}

In this study, we propose safe ASNG, an optimization method that achieves efficient safe optimization in binary spaces based on adaptive stochastic natural gradient (ASNG) method~\cite{asng} employing a Bernoulli distribution as the search distribution.
Safe ASNG computes a safe region using previously evaluated solutions and maps solutions generated according to the Bernoulli distribution to their nearest neighbors within the safe region.
To compute the safe region, safe ASNG constructs a surrogate model of the safety function based on discrete Walsh functions~\cite{walsh:ga:1980,walsh:surrogate:ppsn} and estimates the Lipschitz constant of the safety function with respect to (w.r.t.) the Hamming distance.
When multiple nearest safe solutions exist within the safe region under the Hamming distance, the mapping destination is determined so as to maximize the corrected likelihood.
In addition, we introduce a constraint-handling mechanism that determines the rankings of solutions by taking the safety function values into account.
Furthermore, the initial distribution parameters are determined using safe seeds, which are safe solutions given at the beginning of the optimization.

In the experimental evaluation, we constructed benchmarks for safe optimization based on binary optimization benchmark functions and performed optimization on them.
As a result, while the comparative methods failed to suppress unsafe solution evaluations, we confirmed that the proposed method achieved efficient optimization while suppressing unsafe solution evaluations.

\section{Related Works}

\subsection{Safe Optimization}

In this study, we modify the formulation of safe optimization in the literature~\cite{survey} by changing the search space to a binary space.
We consider the constrained optimization problem for an objective function $f: \{0,1\}^d \to \R$ and safety functions $s_j: \{0,1\}^d \to \R$ for $j=1, \cdots, p$ as
\begin{align}
    \max_{\x \in \{0,1\}^d} \, f(\x) \quad \text{s.t.} \quad s_j(\x) \geq 0  \enspace \text{for all} \enspace j=1, \cdots, p \enspace.
\end{align}
The objective of safe optimization is to optimize the objective function while avoiding the evaluation of solutions that violate constraints imposed by the safety functions \new{during the optimization process}.
In the following, solutions that satisfy all the safety constraints are referred to as \emph{safe solutions}, whereas solutions that violate at least one safety constraint are referred to as \emph{unsafe solutions}.
In addition, in safe optimization, $N_\mathrm{seed}$ safe solutions, called safe seeds, are provided at the beginning of the optimization.

In this study, we consider optimization problems in which the objective function and the safety functions are evaluated simultaneously.
Both the objective function and the safety functions are assumed to be black-box and noise-free.
For simplicity, we further assume that the values of the safety functions can be obtained even when their constraints are violated. 
We note that these settings are the same as the settings in~\cite{safecmaes}, which considers the continuous search space.

\begin{algorithm}[!t] 
\centering
\caption{ASNG with Bernoulli distribution}
\begin{algorithmic}[1] \label{alg:asng}
\REQUIRE Objective function $f$ to be maximized
\REQUIRE $\theta_{\min} = 1/d, \theta_{\max} = 1 - 1/d, \alpha=1.5, \delta_\mathrm{init}=1$
\STATE Set $\btheta^{(0)} = (0.5, \cdots, 0.5)$, $\boldsymbol{s}^{(0)} = \boldsymbol{0}$, $\gamma^{(0)} = 0$, $\delta^{(0)}=\delta_\mathrm{init}$, $t=0$
\WHILE{termination condition is not met}
\FOR{$i = 1, \cdots, \lambda$}
\STATE Generate $\x^{\langle i \rangle} \sim \mathrm{Bernoulli}(\btheta^{(t)})$.
\STATE Evaluate $f(\x^{\langle i \rangle})$.
\ENDFOR
\STATE Compute the estimated natural gradient $\boldsymbol{G}(\btheta^{(t)})$ as \eqref{eq:asng:ng}.
\STATE Update the distribution parameter as \eqref{eq:asng:update:distribution}.
\STATE Project $\btheta^{(t+1)}$ onto $[\theta_{\min}, \theta_{\max}]^d$.
\STATE Set the accumulation factor as $\beta = \delta^{(t)} / \sqrt{d}$
\STATE Compute $\boldsymbol{s}^{(t+1)} = (1 - \beta) \boldsymbol{s}^{(t)} + \sqrt{\beta (2 - \beta)} \frac{ \mathbf{F}_{\btheta}^{\frac 12} \boldsymbol{G}(\btheta^{(t)}) }{ \| \boldsymbol{G}(\btheta^{(t)}) \|_{\mathbf{F}_{\btheta}} }$.
\STATE Compute $\gamma^{(t+1)} = (1 - \beta)^2 \gamma^{(t)} + \beta (2 - \beta)$.
\STATE Update the learning rate $\delta^{(t)}$ as \eqref{eq:asng:update:lr:1} and \eqref{eq:asng:update:lr:2}.
\STATE $t \leftarrow t + 1$
\ENDWHILE
\end{algorithmic} 
\end{algorithm}
%
%
%

\subsection{Adaptive Stochastic Natural Gradient}

Adaptive stochastic natural gradient (ASNG) method is a probabilistic model-based optimization method with a learning rate adaptation mechanism.
\new{ASNG achieves robust optimization performance without hyperparameter tuning.}
ASNG employs a parametric family of probability distributions $\mathcal{P} = \{ P_{\btheta} : \btheta \in \Theta \}$ on the search space $\mathcal{X}$.
Then, it transforms the maximization problem of the objective function $f$ into the maximization problem of the expected objective function value as
\begin{align}
    J(\btheta) = \int_{\x \in \X} f(\x) p_{\btheta}(\x) \diff \x \enspace,
\end{align}
where $p_{\btheta}$ is the density function of $P_{\btheta}$. 
In this paper, we consider the binary search space $\mathcal{X} = \{0,1\}^d$ and the Bernoulli distribution as the search distribution as
\begin{align}
    p_{\btheta}(\x) = \prod_{i=1}^d (\theta_i)^{x_i} (1 - \theta_i)^{1 - x_i} \enspace.
\end{align}
ASNG updates $\btheta$ along the steepest direction w.r.t. the Kullback-Leibler (KL) divergence, which is given by the natural gradient direction $\tilde{\nabla} J(\btheta) = \mathbf{F}_{\btheta}^{-1} {\nabla} J(\btheta)$~\cite{amari:1998:nc} w.r.t. the Fisher metric defined by the Fisher information matrix $\mathbf{F}_{\btheta}$.
Because the natural gradient cannot be obtained analytically in the black-box optimization scenario, ASNG approximates it by Monte Carlo estimation with $\lambda$ samples $\x^{\langle 1 \rangle}, \cdots, \x^{\langle \lambda \rangle} \sim P_{\btheta^{(t)}}$.
Introducing the ranking-based utility $u: \mathcal{X} \to \R$, the estimated natural gradient for Bernoulli distribution is obtained as
\begin{align}
\boldsymbol{G}(\btheta^{(t)}) = \frac{1}{\lambda} \sum^\lambda_{i=1} u(\x^{\langle i \rangle}) ( \x^{\langle i \rangle} - \btheta^{(t)} ) \enspace.  \label{eq:asng:ng}
\end{align}
Then, ASNG updates the distribution parameters as
\begin{align}
\btheta^{(t+1)} &= \btheta^{(t)} + \epsilon^{(t)} \boldsymbol{G}(\btheta^{(t)}) \enspace, \label{eq:asng:update:distribution}
\end{align}
where $\epsilon^{(t)} = \delta^{(t)} / \| \boldsymbol{G}(\btheta^{(t)}) \|_{\mathbf{F}_{\btheta}} $ is the learning rate controlled by $\delta^{(t)}$.
In addition, to prevent premature convergence, we impose margins on the distribution parameters as
\begin{align}
\theta^{(t+1)}_i \leftarrow \max \left\{ \min \left\{ \theta^{(t+1)}_i, \theta_{\max} \right\}, \theta_{\min} \right\} \enspace \text{for} \enspace i=1,\cdots, d \enspace,
\end{align}
where $\theta_{\min}$ and $\theta_{\max}$ are the lower and upper limits of the distribution parameters, respectively.
In this paper, we set $\lambda = 2$ and $(u(\x^{\langle 1 \rangle}), u(\x^{\langle 2 \rangle})) = (+1, -1)$ when $f(\x^{\langle 1 \rangle}) \geq f(\x^{\langle 2 \rangle})$ and $(-1, +1)$ otherwise.

A key point of ASNG is the adaptation mechanism of the learning rate $\delta^{(t)}$.
ASNG maintains the learning rate proportional to the signal-to-noise ratio (SNR) of the update direction $G(\btheta^{(t)})$ at most.
To estimate the SNR value, ASNG introduces two accumulations as
\begin{align}
\boldsymbol{s}^{(t+1)} &= (1 - \beta) \boldsymbol{s}^{(t)} + \sqrt{\beta (2 - \beta)} \mathbf{F}_{\btheta}^{\frac 12} \boldsymbol{G}(\btheta^{(t)}) \\
\gamma^{(t+1)} &= (1 - \beta)^2 \gamma^{(t)} + \beta (2 - \beta) \| \boldsymbol{G}(\btheta^{(t)}) \|_{\mathbf{F}_{\btheta}}^2 \enspace, 
\end{align}
where $\beta > 0$ is the accumulation factor that satisfies $\beta \propto \delta^{(t)}$.
We note these accumulations use the normalized update direction, as shown in Algorithm~\ref{alg:asng}.
Based on these accumulations, ASNG updates the learning rate as
\begin{align}
\delta^{(t+1)} &= \delta^{(t)} \exp\left( \beta \left( \frac{\| \boldsymbol{s}^{(t+1)} \|^2}{ \alpha } - \gamma^{(t+1)} \right) \right) \label{eq:asng:update:lr:1} \\
\delta^{(t+1)} &\leftarrow \min\{ \delta^{(t+1)}, \delta_\mathrm{init} \} \enspace, \label{eq:asng:update:lr:2}
\end{align}
where $\alpha > 0$ and $\delta_\mathrm{init} > 0$ are hyperparameters.

\subsection{Surrogate Model Using Discrete Walsh Functions}
Walsh functions~\cite{walsh} form a complete orthonormal system for functions defined on the closed interval $[0,1]$, and constitute a function system whose theory is developed based on binary representations.
Each basis function is defined for a natural number represented in binary form $\boldsymbol{\ell} = (\ell_1, \ell_2, \cdots)$, using the fractional binary digits $\hat{x}_1, \hat{x}_2, \cdots$ of the input $x \in [0,1]$, as
\begin{align}
    \varphi_\ell(x) = (-1)^{\sum_{i=1}^\infty \ell_i \hat{x}_i} \enspace.
\end{align}
Each basis function takes values in $\pm 1$, and any Lebesgue-integrable function on the interval $[0,1]$ can be represented as a linear combination of these basis functions.

Based on Walsh functions, the discrete Walsh transform, which provides an orthonormal basis for pseudo-Boolean functions, has been proposed~\cite{walsh:ga:1980}.
In the discrete Walsh transform, we consider the discrete Walsh functions with a $d$-dimensional binary vector $\x = (x_1, \cdots, x_d)$ as input, which is defined as
\begin{align}
    \varphi_\ell(\x) = (-1)^{\sum_{i=1}^d \ell_i x_i} \enspace.
\end{align}
In the discrete Walsh transform, a pseudo-Boolean function is expanded as a weighted sum of discrete Walsh functions.

Based on this framework, surrogate models for binary optimization using discrete Walsh functions have been proposed~\cite{walsh:surrogate:ppsn}.
In this approach, a surrogate model $\hat{g}$ of a function $g$ defined on a binary space is constructed as a linear combination of discrete Walsh functions whose order $o(\varphi_{\ell})$, i.e., the number of ones contained in the binary representation of the natural number $\ell$, is at most $R$, as
\begin{align}
    \hat{g}(\x) = \sum_{\ell \text{ s.t. } o(\varphi_{\ell}) \leq R} \hat{w}_\ell \cdot \varphi_\ell(\x) \enspace,
    \label{eq:walsh:surrogate}
\end{align}
where $\hat{w}_\ell \in \R$ is the coefficient corresponding to $\varphi_\ell$.
It has been experimentally demonstrated that this surrogate model based on discrete Walsh functions can achieve a more accurate function approximation for binary optimization problems than Gaussian process regression and polynomial regression~\cite{wsao:gecco:2019}.

\begin{algorithm}[!t] 
\centering
\caption{Safe ASNG}
\begin{algorithmic}[1] \label{alg:safeasng}
\REQUIRE Objective function $f$ to be maximized
\REQUIRE Safety functions $s_1, \cdots, s_p$ 
\REQUIRE Safe seeds $\x_{\mathrm{seed}}^{\langle 1 \rangle}, \cdots, \x_{\mathrm{seed}}^{\langle N_\mathrm{seed} \rangle}$
\REQUIRE $N_\mathrm{safe} = 10 \times d, T_\mathrm{data}=10 \times d, \zeta_\mathrm{data} = 10$
\REQUIRE $\theta_{\min} = 1/d, \theta_{\max} = 1 - 1/d, \alpha=1.5, \delta^{(0)}=1$
\STATE Set $t=0$ and $\mathcal{A} = \{ \x_{\mathrm{seed}}^{\langle k \rangle} \}_{k=1}^{N_\mathrm{seed}}$
\STATE Set $\btheta^{(0)}$ as (\ref{eq:proposed:init:1}) and (\ref{eq:proposed:init:2}) using safe seeds
\WHILE{termination condition is not met}
\STATE Construct surrogate models $\hat{s}_1, \ldots \hat{s}_p$ for $s_1, \ldots s_p$ with $\mathcal{A}$.
\STATE Estimate Lipschitz constants $\hat{L}_1, \cdots, \hat{L}_p$.
\STATE Select the latest $N_\mathrm{safe}$ solutions $\mathcal{D}$ in $\mathcal{A}$ with safety function values no less than their Lipschitz constants.
\FOR{$i = 1, \cdots, \lambda$}
\STATE Generate $\x^{\langle i \rangle} \sim \mathrm{Bernoulli}(\btheta^{(t)})$.
\STATE Project $\x^{\langle i \rangle}$ onto the safe region using $\mathcal{D}$.
\IF{$i > 1$}
\STATE Repair $\x^{\langle i \rangle}$ using other samples $\x^{\langle 1 \rangle}, \cdots, \x^{\langle i-1 \rangle}$.
\ENDIF
\STATE Evaluate $f(\x^{\langle i \rangle})$ and $s_1(\x^{\langle i \rangle}), \cdots, s_p(\x^{\langle i \rangle})$.
\ENDFOR
\STATE Compute the estimated natural gradient $\boldsymbol{G}(\btheta^{(t)})$ as \eqref{eq:asng:ng}.
\STATE Update the distribution parameter $\btheta^{(t+1)}$ and learning rate $\delta^{(t)}$ using the update procedure of ASNG.
\STATE Add evaluated samples to $\mathcal{A}$ if they are not contained.
\STATE $t \leftarrow t + 1$
\ENDWHILE
\end{algorithmic} 
\end{algorithm}
%
%
%

\section{Proposed Method: Safe ASNG}

In this paper, we propose safe ASNG, an optimization method for safe optimization in binary optimization problems.
Safe ASNG constructs a surrogate model of the safety functions based on discrete Walsh functions and estimates their Lipschitz constants w.r.t. the Hamming distance.
Then, based on the estimated Lipschitz constants, safe ASNG computes a safe region centered at previously evaluated safe solutions.
Safe ASNG projects generated solutions to the closest point with the largest likelihood in this region so as to suppress unsafe solution evaluations.
The pseudocode of the proposed method is shown in Algorithm~\ref{alg:safeasng}.

\subsection{Estimation of Lipschitz Constant}

Safe ASNG constructs surrogate models $\hat{s}_1, \cdots, \hat{s}_p$ for each safety function using an archive $\mathcal{A}$ that stores all evaluated solutions and their evaluation values.
In existing work, Gaussian process regressions are used as surrogate models of safety functions in continuous spaces~\cite{safecmaes}.
Because Gaussian process regression requires a computational cost of $O((N_\mathrm{data})^3)$ w.r.t. the number of training data points $N_\mathrm{data}$, training data are limited to the samples evaluated in the five most recent iterations in \cite{safecmaes}.
However, unlike in the continuous space, the search distribution in ASNG with Bernoulli distributions covers the whole of the search space, especially when \del{applying the margin correction}{}\new{imposing the margin}.
As a result, limiting the training data to only a few recent samples leads to an excessively small training set, which is undesirable for constructing reliable surrogate models. 

\begin{figure}[t]
    \centering
    \includegraphics[width=0.99\linewidth]{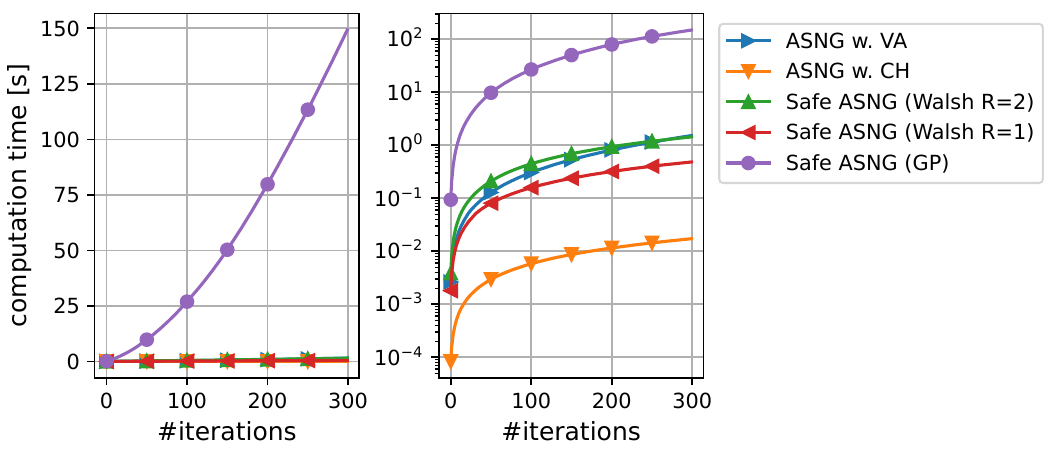} \\
    \caption{The cumulative computational time required for the update procedure on a 10-dimensional problem (Left: linear scale, Right: log scale).\protect\footnotemark[1]
    We plot the average of five trials on the objective function that returns a random value and a safety function that returns a constant positive value.}
    \label{fig:time}
\end{figure}

Therefore, we use discrete Walsh-based surrogates as computationally efficient surrogate models for safety functions.
Figure~\ref{fig:time} shows the computational time required for updates of safe ASNG when Gaussian process regressions and discrete Walsh-based surrogates are trained using all evaluated solutions.
We can see that the discrete Walsh-based surrogates reduce the computational time to around one-hundredth of that with Gaussian process regressions.
Furthermore, it has been experimentally demonstrated that surrogate models based on discrete Walsh functions achieve a more accurate function approximation in binary spaces than Gaussian process regression and polynomial regression~\cite{wsao:gecco:2019,walsh:surrogate:ppsn}.
The surrogate coefficients $\hat{w}_\ell$ are computed using the least squares method. 

Next, for each surrogate model $\hat{s}_j$, the Lipschitz constant w.r.t. the Hamming distance is estimated as follows.
First, we generate $100$ samples $\mathcal{S}_\mathrm{c}$ from the current probability distribution $P_{\btheta^{(t)}}$.
Then, for each generated sample $\x_\mathrm{c}$ in $\mathcal{S}_\mathrm{c}$, we compute neighboring solutions $\mathcal{S}_\mathrm{n}(\x_\mathrm{c})$ whose Hamming distance from the generated sample is one and evaluate them using the surrogate model.
Based on these evaluations, the Lipschitz constant (for the local region with high generation probability) is estimated as their maximum difference
\begin{align}
    \hat{L}_j' = \max_{\x_\mathrm{c} \in \mathcal{S}_\mathrm{c}} \max_{\x_\mathrm{n} \in \mathcal{S}_\mathrm{n}(\x_\mathrm{c})} \left| \hat{s}_j(\x_\mathrm{c}) - \hat{s}_j(\x_\mathrm{n}) \right| \enspace.
\end{align}
\new{We note that the Lipschitz constant is slightly overestimated due to the model's prediction error.}

\footnotetext[1]{The computational time is measured using an Intel Xeon w7-2595X with \texttt{NumPy\,1.26.4} and \texttt{scikit-learn\,1.5.2}. We used \texttt{GaussianProcessRegressor} and \texttt{LinearRegression} in \texttt{scikit-learn} for Gaussian process regression and discrete Walsh-based surrogate, respectively.}

\paragraph{Overestimation under Small Training Dataset}

Following~\cite{safecmaes}, we increase the estimated Lipschitz constant when the number of training data $N_\mathrm{data} := |\mathcal{A}|$ for the surrogate models is small.
Specifically, we increase the Lipschitz constant when $N_\mathrm{data} < T_\mathrm{data}$ as
\begin{align}
    \hat{L}_j' \leftarrow \hat{L}_j' \cdot (\zeta_\mathrm{data})^{1 - N_\mathrm{data} / T_\mathrm{data}} \enspace,
\end{align}
where $T_\mathrm{data} = 10 \times d$ and $\zeta_\mathrm{data} = 10$ are hyperparameters.

\subsection{Computation of Safe Region}

The safe region is defined as a region in which all solutions are safe.
If a safety function $s_j$ has a Lipschitz constant $L_j$, then for a safe solution $\x_\mathrm{safe}$ and any solution $\x \in \{0,1\}^d$, the following relation holds:
\begin{align*}
    \mathrm{dist}(\x, \x_\mathrm{safe}) &\leq \frac{s_j(\x_\mathrm{safe})}{ L_j }
    \enspace \Rightarrow \enspace
    s_j(\x) \geq 0 \enspace.
\end{align*}
In this paper, the distance function is defined as the Hamming distance.
Based on this relation, safe ASNG considers the union of safe regions centered at previously evaluated solutions, and suppresses unsafe solution evaluations by projecting candidate solutions onto this region.
We construct an archive $\mathcal{D}$ that stores the most recently evaluated $N_\mathrm{safe}$ solutions in the archive $\mathcal{A}$ with the safety function value $s_j(\x_\mathrm{safe})$ no less than $\hat{L}_j$ for all $j=1,\cdots,p$ together with their safety function values. Then, using the archive $\mathcal{D}$, we estimate the safe region around previously evaluated safe solutions as
\begin{multline*}
    \mathcal{X}_\mathrm{safe} = \bigcup_{\x_\mathrm{safe} \in \mathcal{D}} \biggl\{ \x \in \{0,1\}^d 
    \mid \mathrm{dist}(\x, \x_\mathrm{safe}) \leq \min_{j=1,\cdots,p} \frac{s_j(\x_\mathrm{safe})}{ \hat{L}_j} \biggr\} \enspace.
\end{multline*}

\paragraph{Degeneration Handle of Safe Region}
In binary spaces, unlike continuous spaces, if the estimated Lipschitz constant is excessively large, the safe region computed around the evaluated solutions $\mathcal{D}$ may not expand beyond $\mathcal{D}$.
Therefore, we construct another archive $\mathcal{D}_{0}$ that stores the most recently evaluated $N_\mathrm{safe}$ solutions with positive safety function values and correct the estimated Lipschitz constant by using the solution in $\mathcal{D}_{0}$ with the maximum safety function value as
\begin{align}
    \hat{L}_j = \min\left\{ \hat{L}_j', \max_{\x_\mathrm{safe} \in \mathcal{D}_{0}} {s}_j(\x_\mathrm{safe}) \right\}
    \label{eq:lip:mod}
    \enspace.
\end{align}
This correction ensures that the constructed safe region $\mathcal{X}_\mathrm{safe}$ includes the neighborhoods of at least one solution in $\mathcal{D}$, which prevents premature convergence.

\subsection{Projection to Safe Region}

\paragraph{Likelihood-based Projection}
In the solution generation process, safe ASNG generates solutions from the Bernoulli distribution and projects them to the nearest point in the safe region to suppress unsafe solution evaluations.
Unlike the projection to the safe region in continuous space~\cite{safecmaes}, there may exist multiple nearest points within the safe region in binary space.
Therefore, safe ASNG determines the projection destination among these solutions based on the likelihood under the current probability distribution.

First, safe ASNG computes the distance between the solution $\x \sim P_{\btheta^{(t)}}$ before projection and the safe regions centered at solutions in the archive $\mathcal{D}$, and obtains the nearest safe solution $\x_\mathrm{near} \in \mathcal{D}$ as
\begin{align}
    \x_\mathrm{near} &= \argmin_{\x_\mathrm{safe} \in \mathcal{D}} \left\{ \Delta(\x, \x_\mathrm{safe}) \right\} \enspace,
\end{align}
where $\Delta(\x, \x_\mathrm{safe})$ represents the signed distance from $\x$ to the boundary of the safe region centered at $\x_\mathrm{safe}$ and is defined as
\begin{align*}
    \Delta(\x, \x_\mathrm{safe}) = \mathrm{dist}(\x, \x_\mathrm{safe}) - \min_{j=1,\cdots,p} \frac{s_j(\x_\mathrm{safe})}{\hat{L}_j} \enspace.
\end{align*}
If there exist multiple nearest safe solutions, we select the one that was evaluated most recently.

If the solution $\x$ before projection lies in the safe region centered at $\x_\mathrm{near}$, i.e., if $\Delta(\x, \x_\mathrm{near})$ is non-positive, no projection is performed.
Otherwise, for each bit $i$, we compute the increase in likelihood after flipping the $i$-th bit as
\begin{align*}
    r(i;\x) = \begin{cases}
    p_{\btheta^{(t)}}(\bar{\x}(i)) - p_{\btheta^{(t)}}(\x) & \text{if} \enspace x_i \neq x_{\mathrm{near}, i} \\
    - \infty & \text{if} \enspace x_i = x_{\mathrm{near}, i} \enspace,
    \end{cases}
\end{align*}
where $\bar{\x}(i)$ denotes the binary vector obtained by flipping only the $i$-th bit of $\x$, i.e., $\bar{\x}(i) = (x_1, \cdots, x_{i-1}, 1 - x_i, x_{i+1}, \cdots, x_d)$.
Then, we flip the $n_\mathrm{flip} = \lceil \Delta(\x, \x_\mathrm{near}) \rceil$ bits with the largest increases $r(i;\x)$ so as to move $\x$ closer to the safe solution $\x_\mathrm{near}$.
\new{We consider that this likelihood-based selection is reasonable when the distribution parameter is converging.}

\paragraph{Generation of Distinct Samples}

Safe ASNG generates $\lambda$ solutions in each iteration; however, if these $\lambda$ solutions are projected onto the same solution in the safe region, the probability distribution parameters may not change, which can cause stagnation of the optimization.
To mitigate this issue, if the projected solution has already been generated in the current iteration, we project the solution to another destination that yields the next-largest likelihood increase.
Specifically, when $\lambda=2$, we determine the projection destination by re-flipping the dimension with the $n_\mathrm{flip}$-th largest value of $r(i;\x)$ and instead flipping the dimension with the $(n_\mathrm{flip}+1)$-th largest value.

\subsection{Ranking-based Constraint Handling}
\label{sec:proposed:constrant-handling}

When the representative ability of the surrogate models is insufficient to fit the safety functions, the estimated safe region can contain unsafe solutions due to underestimation of the Lipschitz constants, which may cause unsafe evaluations.
Therefore, we introduce a constraint-handling mechanism as an additional component to suppress unsafe solution evaluations, which was introduced in~\cite{Deb:2003:constraint}.
In this mechanism, the preference relation between two generated solutions $\x$ and $\x'$ is determined as follows, where $\x \preccurlyeq_{f,s} \x'$
indicates that $\x'$ is preferred to or considered no worse than $\x$:
\begin{multline*}
    \x \preccurlyeq_{f,s} \x' \Leftrightarrow 
    \begin{cases}
        f(\x) \leq f(\x') & \begin{array}{l}
        \text{if} \enspace \mathop{\rm min}\limits_{\y \in \{ \x, \x'\}} \{s_{<0}(\y)\} \geq 0 \\
        \text{or} \enspace  s_{<0}(\x) = s_{<0}(\x')
        \end{array} \\
        s_{<0}(\x) < s_{<0}(\x') & \enspace\text{otherwise} \enspace ,
    \end{cases}
\end{multline*}
where $s_{<0}(\x)$ denotes the amount of safety violation, defined as
\begin{align}
    s_{<0}(\x) = \sum_{j=1}^p \min\{ s_j(\x), 0 \} \enspace.
\end{align}
This preference relation is designed so that safe solutions are always preferred to unsafe solutions.
Comparisons between safe solutions are performed based on the objective function values, whereas comparisons between unsafe solutions are performed based on the amount of constraint violation.

\subsection{Initialization with Safe Seeds}
\label{sec:proposed:init}

In safe ASNG, the probability distribution parameters are initialized using the $N_\mathrm{seed}$ safe seeds $\x_{\mathrm{seed}}^{\langle 1 \rangle}, \cdots, \x_{\mathrm{seed}}^{\langle N_\mathrm{seed} \rangle}$ provided at the beginning of the optimization as 
\begin{align}
    \btheta^{(0)} &= \frac{1}{N_\mathrm{seed}} \sum_{k=1}^{N_\mathrm{seed}} \x_{\mathrm{seed}}^{\langle k \rangle}
    \enspace.
    \label{eq:proposed:init:1} 
\end{align}
In addition, for each dimension $i$, the margin \del{correction is applied}{}\new{is imposed} to project the parameter onto $[\theta_{\min}, \theta_{\max}]$ as
\begin{align}
    \theta^{(0)}_i \leftarrow \max \left\{ \min \left\{ \theta^{(0)}_i, \theta_{\max} \right\}, \theta_{\min} \right\} \enspace.
    \label{eq:proposed:init:2}
\end{align}


%
%
%
\begin{figure*}[t]
    \centering

    \includegraphics[width=0.32\linewidth]{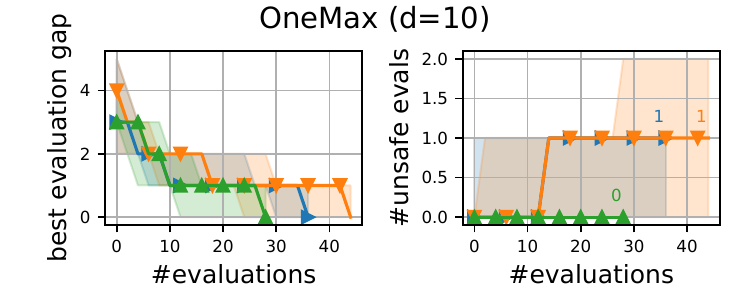}
    \includegraphics[width=0.32\linewidth]{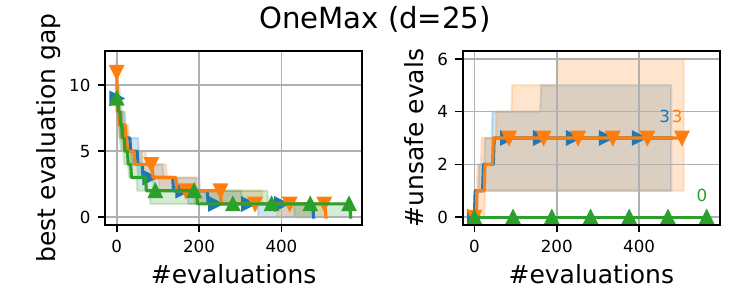}
    \includegraphics[width=0.32\linewidth]{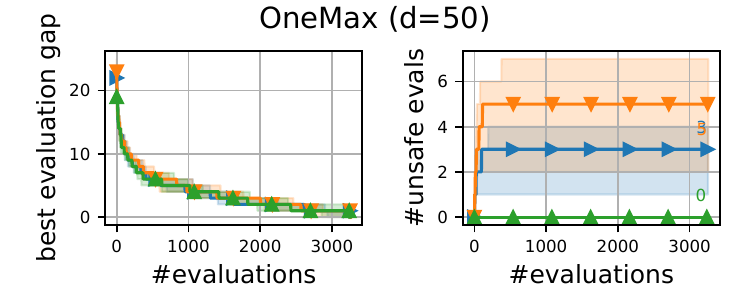}
    
    \includegraphics[width=0.32\linewidth]{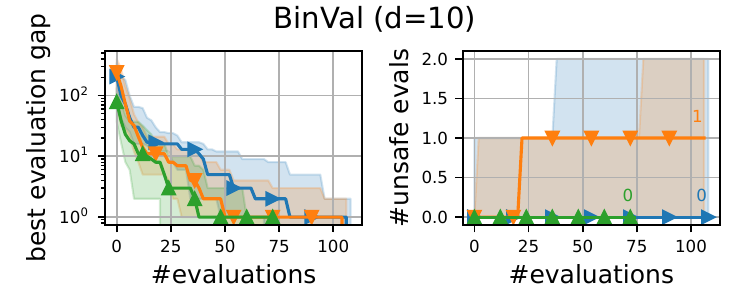}
    \includegraphics[width=0.32\linewidth]{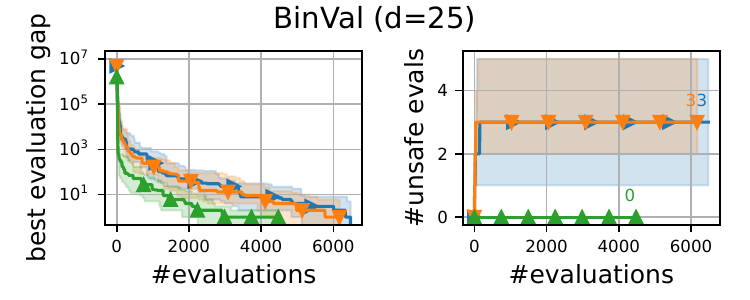}
    \includegraphics[width=0.32\linewidth]{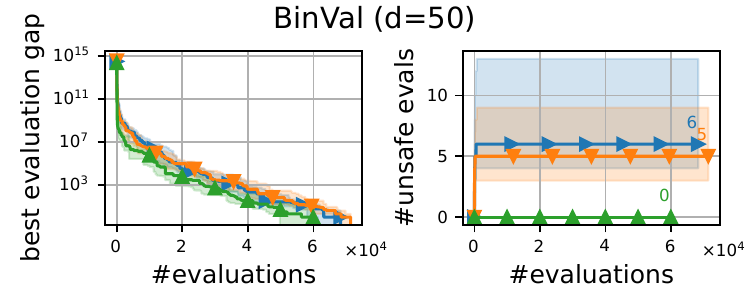}
    
    \includegraphics[width=0.32\linewidth]{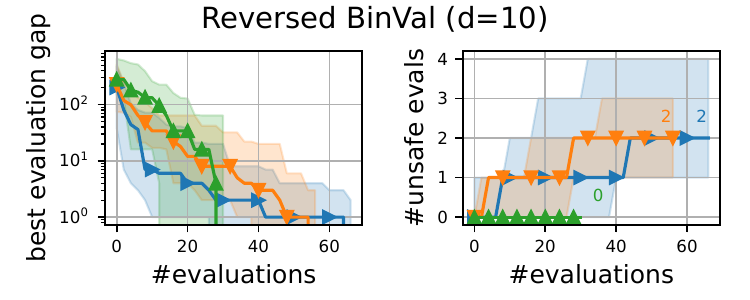}
    \includegraphics[width=0.32\linewidth]{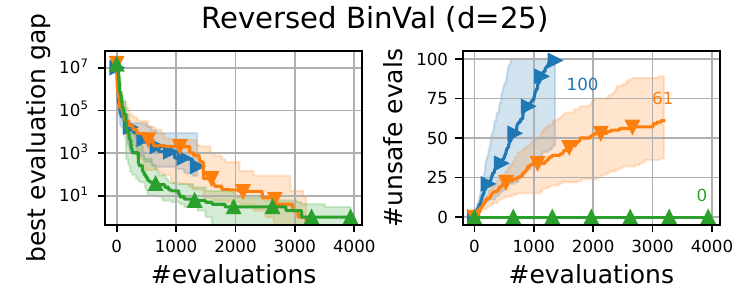} 
    \includegraphics[width=0.32\linewidth]{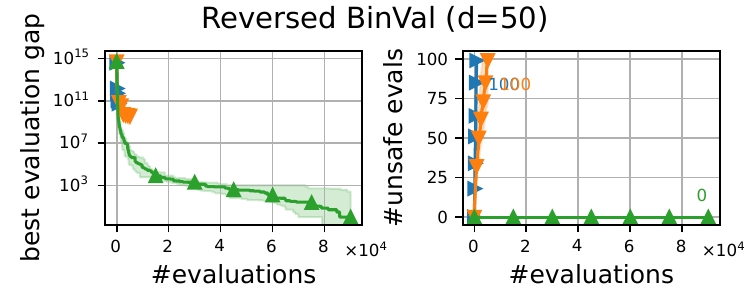} 
        
    \includegraphics[width=0.32\linewidth]{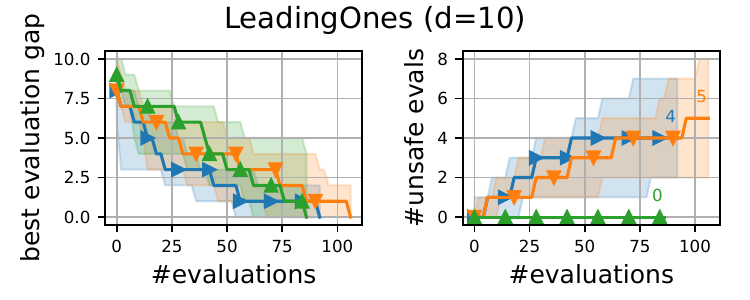} 
    \includegraphics[width=0.32\linewidth]{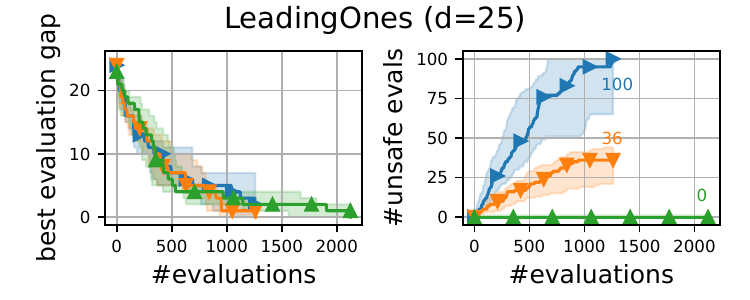} 
    \includegraphics[width=0.32\linewidth]{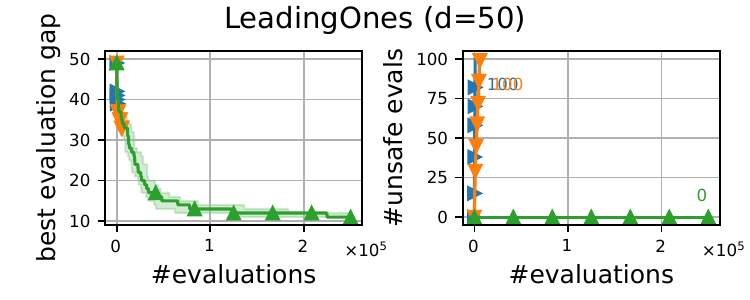} 
    
    \vspace*{0.2cm} \includegraphics[width=0.3\linewidth]{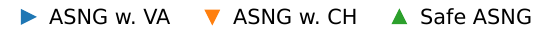} \\
    \caption{Result on Experiment~1, compatible settings. We plot the transitions of the best evaluation gap, which is the difference between the optimal function value and the best evaluation value (in the left figure), and the number of unsafe evaluations (in the right figure). These plots show the median and interquartile ranges over 25 trials. Additionally, in the right figure, we write the median number of unsafe evaluations at the end of the optimization. }
    \label{fig:exp1}
\end{figure*}
\begin{figure*}[t]
    \centering

    \includegraphics[width=0.32\linewidth]{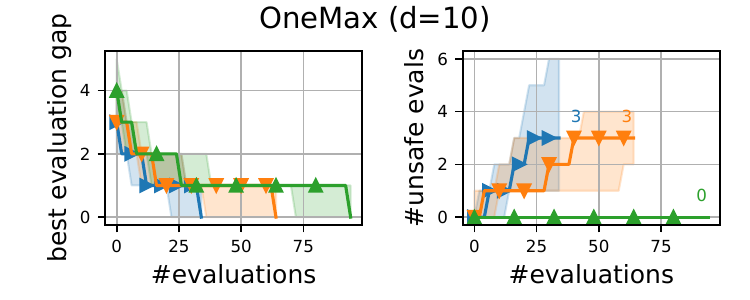}
    \includegraphics[width=0.32\linewidth]{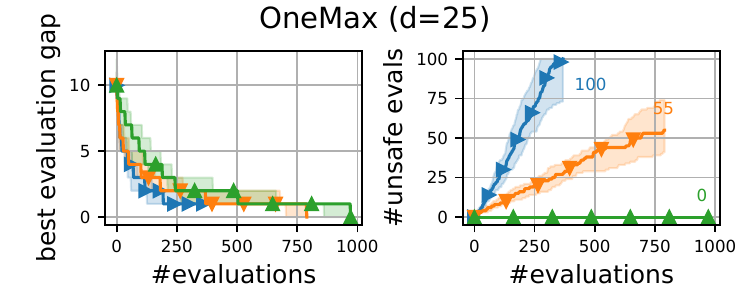}
    \includegraphics[width=0.32\linewidth]{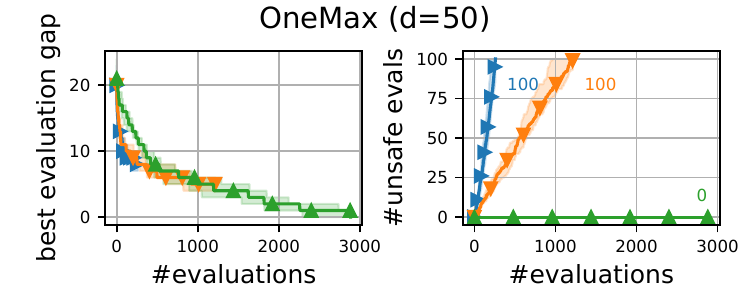}
    
    \includegraphics[width=0.32\linewidth]{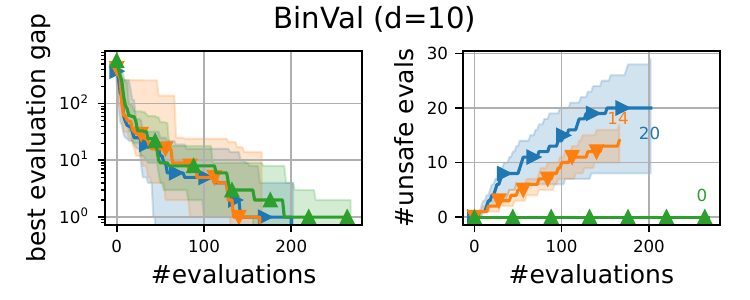} 
    \includegraphics[width=0.32\linewidth]{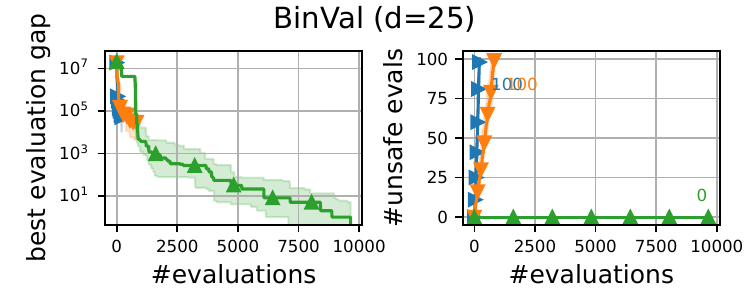} 
    \includegraphics[width=0.32\linewidth]{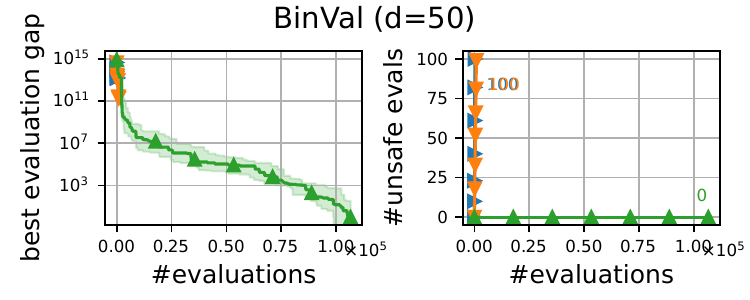}
    
    \includegraphics[width=0.32\linewidth]{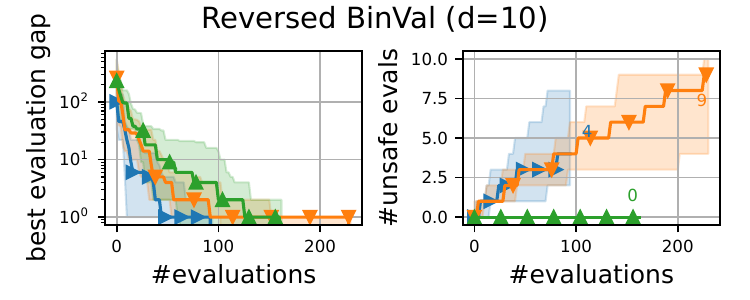} 
    \includegraphics[width=0.32\linewidth]{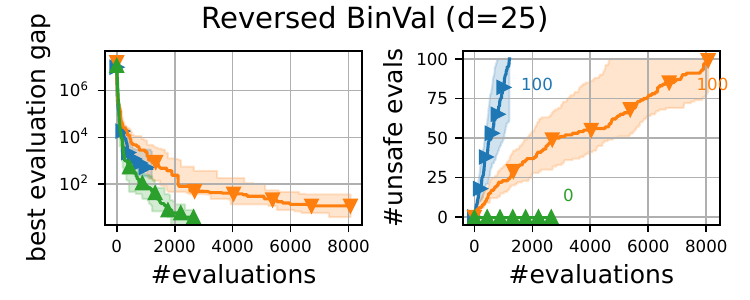} 
    \includegraphics[width=0.32\linewidth]{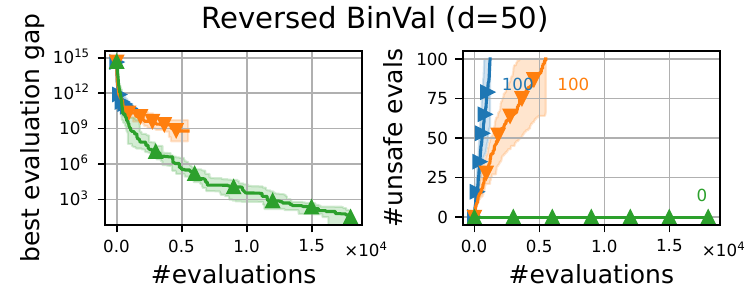} 
        
    \includegraphics[width=0.32\linewidth]{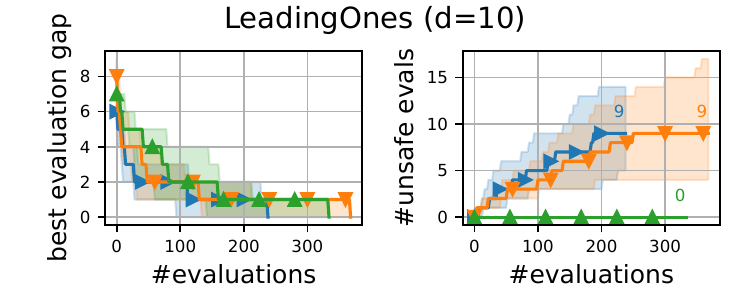} 
    \includegraphics[width=0.32\linewidth]{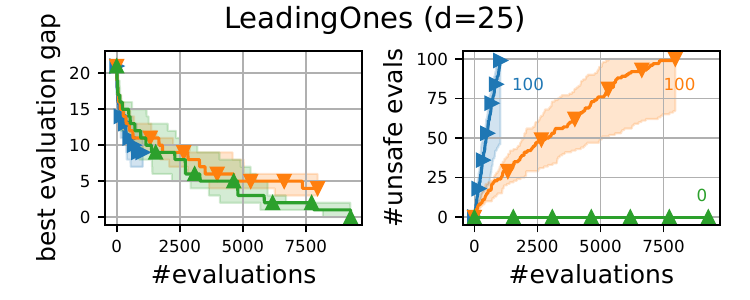} 
    \includegraphics[width=0.32\linewidth]{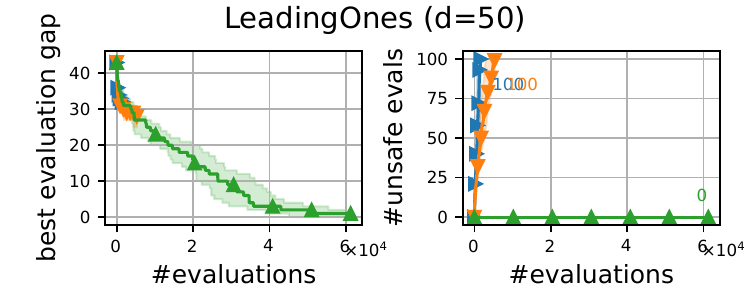} 
    
    \vspace*{0.2cm} \includegraphics[width=0.3\linewidth]{image_exp1/label.pdf} \\
    \caption{Result on Experiment~2, conflicting settings. We plot the transitions of the best evaluation gap, which is the difference between the optimal function value and the best evaluation value (in the left figure), and the number of unsafe evaluations (in the right figure). These plots show the median and interquartile ranges over 25 trials. Additionally, in the right figure, we write the median number of unsafe evaluations at the end of the optimization. }
    \label{fig:exp2}
\end{figure*}
%
%
%

\section{Experiment}

In the experimental evaluation, we assessed the performance of the proposed method in two settings as follows:
\begin{itemize}
    \item Experiment~1: objective and safety functions are compatible.
    \item Experiment~2: objective and safety functions are conflicting.
\end{itemize}


\subsection{Comparison Methods}

\paragraph{ASNG with Violation Avoidance (ASNG w. VA)}
In the experimental evaluation, we used ASNG equipped with violation avoidance~\cite{Kaji:2009}, which is a generic constraint-handling method applicable to evolutionary computation methods for safe optimization, as one of the comparative methods.
In violation avoidance, when generating a solution, if the nearest previously evaluated solution is an unsafe solution, the solution generation process is retried so as to avoid unsafe solution evaluations.
The distance between a newly generated solution $\x_\mathrm{new}$ and a previously evaluated solution $\x_\mathrm{old}$ is defined as
\begin{align}
    d_{\mathrm{VA}}(\x_\mathrm{new}, \x_\mathrm{old}) = \frac{\mathrm{dist}( \x_\mathrm{new}, \x_\mathrm{old} )}{w(\x_\mathrm{old})} \enspace,
\end{align}
where the weight $w(\x_\mathrm{old}) \in \R_{>0}$ is determined based on whether $\x_\mathrm{old}$ is safe or unsafe as
\begin{align}
    w(\x_\mathrm{old}) = \begin{cases}
    w_\mathrm{safe} & \text{if $\x_\mathrm{old}$ is safe} \\ 
    w_\mathrm{unsafe} & \text{if $\x_\mathrm{old}$ is unsafe} \enspace .
    \end{cases}
\end{align}
With this distance function, a larger weight $w_\mathrm{safe}$ makes it more likely to generate new solutions in the neighborhood of safe solutions, while a larger weight $w_\mathrm{unsafe}$ makes it more likely to generate solutions farther away from safe solutions previously evaluated.

In the experiments, we set $w_\mathrm{safe} = w_\mathrm{unsafe} = 1$.
In the generation process of each solution, we generated $10 \times d$ samples from the Bernoulli distribution and randomly selected one sample whose nearest evaluated solution is safe.
If multiple nearest solutions existed, we used the median of their safety function values and regarded the constraints as satisfied when the median was non-negative.
If no sample in $10 \times d$ samples satisfied the condition, the optimization was terminated.
In addition, we initialized the distribution parameters according to the initialization method described in Section~\ref{sec:proposed:init}.

\paragraph{ASNG with Constraint Handling (ASNG w. CH)}
In addition, we employed another comparative method, ASNG with the constraint-handling mechanism introduced in Section~\ref{sec:proposed:constrant-handling} and the initialization method described in Section~\ref{sec:proposed:init}.

\subsection{Experimental Setting}

We used the following benchmark functions for binary optimization as objective functions:
\begin{itemize}
    \item OneMax: $f(\x) = \sum_{i=1}^d x_i$
    \item LeadingOnes: $f(\x) = \sum_{i=1}^d \prod_{j=1}^i x_j$
    \item BinVal: $f(\x) = \sum_{i=1}^d 2^{i-1} x_i$
    \item Reversed BinVal: $f(\x) = \sum_{i=1}^d 2^{d-i} x_i$
\end{itemize}
OneMax, LeadingOnes, and BinVal are commonly-used benchmark functions on the binary domain. OneMax evaluates the number of ones in a binary string, LeadingOnes evaluates the number of consecutive ones from the beginning of the binary string, and BinVal evaluates the value obtained by interpreting the binary string as a binary number. Reversed BinVal is defined by reversing the coefficients of BinVal, where the leading bits have a larger impact than the trailing ones.

In addition, we provided $N_\mathrm{seed} = 10$ safe solutions as safe seeds, which were selected from uniformly sampled solutions in the search space such that their safety function values are non-negative.
These safe seeds were shared across all methods.
We set the number of dimensions to $d=10,25,50$ and the maximum number of iterations to $d^3$.
The maximum order of the surrogate model based on discrete Walsh functions was set to $R=2$ for $d=10, 25$ and $R=1$ for $d=50$.\footnote{Our implementation of discrete Walsh functions is based on \url{https://gitlab.com/florianlprt/wsao}}
The sample size is set to $\lambda=2$ for all methods.
We terminated the optimization when the number of unsafe evaluations reached 100.
We conducted 25 independent trials for each setting.

\subsection{Result of Experiment~1 (Compatible Setting)}

First, we conducted experiments using the following safety function as
\begin{align}
    s(\boldsymbol{x}) = \left( \sum_{i= \lfloor d/2 \rfloor + 1}^d x_i \right) - \left\lfloor \frac{d}{4} \right\rfloor \enspace.
    \label{eq:constraint1}
\end{align}
The first term of this safety function represents the number of ones in the last $\lfloor d/2 \rfloor$ bits of the binary string, and it is required to keep this number at least $\lfloor d/4 \rfloor$.
For all benchmark functions used in this study, the optimal solution in unconstrained optimization is given by $\x^\ast = (1, \cdots, 1)^\T$, and thus the safety function becomes non-negative in the neighborhood of the optimum.
Therefore, with this safety function, improving the objective function value and improving safety are compatible, and safe solutions are more likely to be generated as the search approaches the optimum.

Figure~\ref{fig:exp1} shows the experimental results.
Each plot reports the transition of the best evaluation gap (i.e., the gap between the optimal value and the best objective value achieved with safe evaluation) on the left side and the transition of the number of unsafe evaluations on the right side.
Focusing on the results for OneMax and BinVal, the improvement in the objective value was comparable to that of the comparison methods, while the number of unsafe evaluations was significantly reduced.
Specifically, the proposed method did not incur unsafe evaluations, whereas the comparison methods evaluated unsafe solutions in the early stage of the optimization process.

Next, we consider the results for Reversed BinVal.
As in the case of BinVal, the proposed method successfully suppressed unsafe evaluations, while the comparison methods failed it.
In the 25- and 50-dimensional problems, the optimization processes of the comparison methods were terminated because the number of unsafe evaluations reached the maximum budget.
In contrast to BinVal, the leading bits have larger impacts on Reversed BinVal, and setting the leading bits to one leads to a rapid improvement of the objective value.
On the other hand, the safety constraint in~\eqref{eq:constraint1} is imposed so that the trailing bits remain one, which prevents the comparison methods from achieving safe optimization.

Finally, we focus on the results for LeadingOnes.
In the 10- and 25-dimensional problems, the number of unsafe evaluations increases throughout the optimization process of the comparison methods.
This was because, for LeadingOnes, these methods optimized the bits from the leading positions, and the distribution parameters for the trailing bits were not sufficiently optimized until the leading positions were optimized.
In addition, the comparison methods failed to achieve safe optimization in the 50-dimensional problem.
In contrast, the proposed method did not incur unsafe evaluations in all cases.
However, because the maximum budget of iterations was reached, the optimization was terminated before the optimal solution was found in the 50-dimensional problem.

\begin{figure*}[t]
    \centering
    \includegraphics[width=0.32\linewidth]{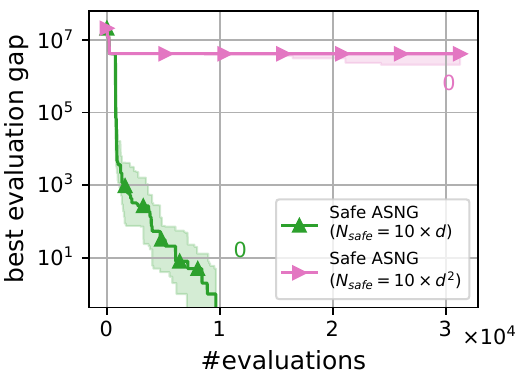} 
    \includegraphics[width=0.65\linewidth]{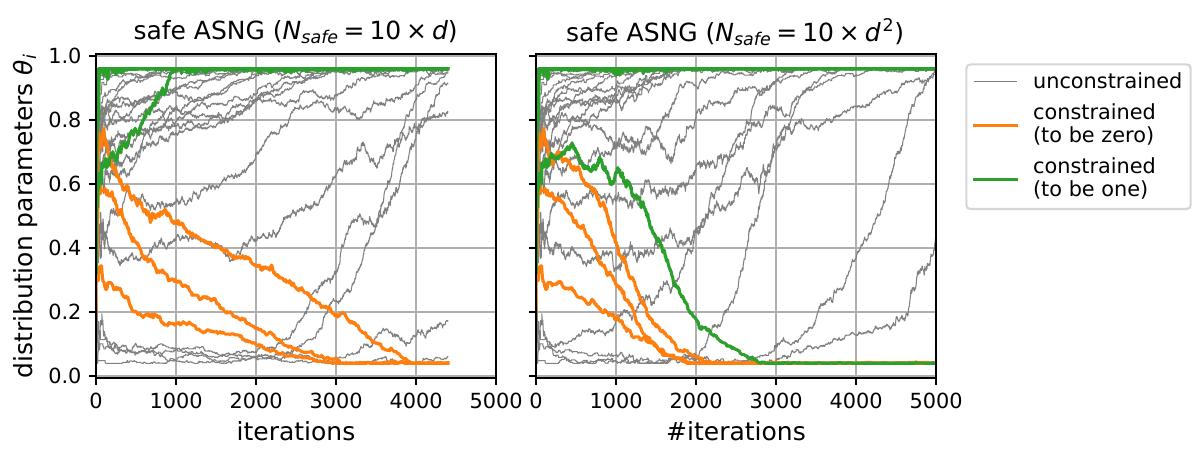}
    \caption{Result of different settings of $N_\mathrm{safe} = 10 \times d$ (recommended setting) and $N_\mathrm{safe} = 10 \times d^2$. We run safe ASNG with $R=2$ on 25-dimensional Binval with the conflicting settings in Experiment~2. Left figure shows the median and interquartile ranges of the best evaluation gap over 25 trials, in addition to the median number of unsafe evaluations at the end of the optimization. Right figures show the transitions of the distribution parameters in typical trials.}
    \label{fig:exp_param_move}
\end{figure*}

\subsection{Result of Experiment~2 (Conflicting Setting)}

Next, we conducted experiments using the following safety function as
\begin{align}
    s(\boldsymbol{x}) = \left\lfloor \frac{d}{8} \right\rfloor - \left( \sum_{i=1}^{\lfloor d/4 \rfloor} x_{d-i+1} \right) \enspace.
    \label{eq:constraint2}
\end{align}
The second term of this safety function represents the number of ones in the last $\lfloor d/4 \rfloor$ bits of the binary string, and it is required to keep this number no greater than $\lfloor d/8 \rfloor$.
That is, it is required to restrict the number of ones in the trailing bits, in which improving the objective value and improving safety are in a conflicting relationship, and unsafe solutions are more likely to be generated as the search approaches the optimum.

Figure~\ref{fig:exp2} shows the experimental results.
Each plot reports the transition of the gap between the optimal value and the best objective value, as well as the transition of the number of unsafe evaluations.
Focusing first on the results for OneMax, as in Experiment~1, only the safe ASNG achieved safe optimization. 
We also observed that, in the 50-dimensional case, the decreasing rate of the best evaluation gap was slow compared to those of comparison methods in the early phase of the optimization.
We consider that this was because the Lipschitz constants were overestimated due to the lack of training data for the surrogate models, and the two sampled solutions were often projected onto nearby points due to the projection mechanism to the safe region, resulting in smaller changes in the probability distribution parameters.

We observed a similar trend for other functions, BinVal, Reversed BinVal, and LeadingOnes: the proposed method successfully found the optimal safe solution while the comparison methods reached the maximum number of unsafe evaluations.
Comparing the result in BinVal with that in Reversed BinVal, the ASNG equipped with the constraint-handling found a better evaluation value in Reversed BinVal than in BinVal because Reversed BinVal has larger coefficients on the leading bits that are not restricted by the safety constraint~\eqref{eq:constraint2}.
We also observed that safe ASNG found the best safe solution in the maximum evaluation budget in 50-dimensional LeadingOnes, while it failed with the compatible setting in Experiment~1, because the optimal solution for LeadingOnes with the safety constraint~\eqref{eq:constraint2} is given by $\x^\ast = (1,\cdots,1,0\cdots,0)$ that contains $\lfloor d/4 \rfloor - \lfloor d/8 \rfloor$ zeros on the trailing part.

\paragraph{Discussion on Hyperparameter Sensitivity}

We investigated the sensitivity of the hyperparameter $N_\mathrm{safe}$, the number of safe solutions used to construct the safe region.
We evaluated safe ASNG with $N_\mathrm{safe} = 10 \times d$ (recommended setting) and $N_\mathrm{safe} = 10 \times d^2$ on 25-dimensional BinVal, \new{which are difficult problems in the conflicting setting}. 
Figure~\ref{fig:exp_param_move} shows the transitions of the best evaluation gap and distribution parameters.
We observed that the best evaluation gap stagnated with $N_\mathrm{safe} = 10 \times d^2$.
From the transitions of the distribution parameters, we observed that some parameters that should converge to one were updated toward zero, which forced the trailing bits to be zero more than necessary.
This was because an overestimation of the Lipschitz constant prevented the safe region from reaching the constraint boundary.
For the recommended setting $N_\mathrm{safe} = 10 \times d$, on the other hand, the modification of the Lipschitz constant in~\eqref{eq:lip:mod} successfully prevented such overestimation.

\section{Conclusion}

In this study, we proposed safe ASNG, an optimization method for performing efficient safe optimization in binary search spaces.
Safe ASNG constructs surrogate models of the safety functions based on discrete Walsh functions, and builds a safe region centered at previously evaluated solutions by estimating the Lipschitz constants of the safety functions with respect to the Hamming distance.
It then suppresses unsafe solution evaluations by projecting solutions sampled from a Bernoulli distribution to their nearest neighbors within the safe region.
In addition, we introduced a constraint-handling mechanism that determines the preference relation between solutions by taking the safety function values into account, and updates the probability distribution parameters so as to avoid unsafe solution evaluations.
In the experimental evaluation, we assessed the performance on the benchmark functions for binary optimization under two types of safety functions: safety functions that are compatible with improving the objective value and safety functions that exhibit a conflicting relationship with improving the objective value.
The proposed method found the optimal safe solution while suppressing unsafe evaluations in almost all settings.

As future work, an important direction is to improve the optimization performance, especially on LeadingOnes.
One possible approach is to also construct a surrogate model for the objective function.
Another direction is to extend the proposed approach to safe optimization for other types of discrete-variable optimization, such as integer and categorical variables.
The sensitivity analysis of hyperparameters in safe ASNG is also left for future work.

\begin{acks}
This study was partially funded by JSPS KAKENHI (JP23K28156, JP23H00491, and JP24K20857) and JST ACT-X (JPMJAX24C7).
We used generative AI tools to improve the quality of writing.
\end{acks}

\bibliographystyle{ACM-Reference-Format}
\bibliography{sample-base}

\end{document}